\pdfoutput=1

\documentclass[11pt]{article}

\usepackage[]{emnlp2021}

\usepackage{times}
\usepackage{latexsym}

\usepackage{multirow}
\usepackage{graphicx}
\usepackage{booktabs}
\usepackage{amsmath}

\usepackage[T1]{fontenc}

\usepackage[utf8]{inputenc}

\usepackage{microtype}

\title{DialogueCSE: Dialogue-based Contrastive Learning of\\Sentence Embeddings}

\author{
    Che Liu, Rui Wang, Jinghua Liu, Jian Sun, Fei Huang, Luo Si \\
    DAMO Academy, Alibaba Group \\
    \texttt{\{liuche.lc,wr224079,shaohan.ljh,jian.sun,f.huang,luo.si\}@alibaba-inc.com} \\
}

\begin{document}

\maketitle

\begin{abstract}
Learning sentence embeddings from dialogues has drawn increasing attention due to its low annotation cost and high domain adaptability.
Conventional approaches employ the siamese-network for this task, which obtains the sentence embeddings through modeling the context-response semantic relevance by applying a feed-forward network on top of the sentence encoders.
However, as the semantic textual similarity is commonly measured through the element-wise distance metrics (e.g. cosine and L2 distance), such architecture yields a large gap between training and evaluating.
In this paper, we propose DialogueCSE, a dialogue-based contrastive learning approach to tackle this issue.
DialogueCSE first introduces a novel matching-guided embedding (MGE) mechanism, which generates a context-aware embedding for each candidate response embedding (i.e. the context-free embedding) according to the guidance of the multi-turn context-response matching matrices.
Then it pairs each context-aware embedding with its corresponding context-free embedding and finally minimizes the contrastive loss across all pairs.
We evaluate our model on three multi-turn dialogue datasets: the Microsoft Dialogue Corpus, the Jing Dong Dialogue Corpus, and the E-commerce Dialogue Corpus.
Evaluation results show that our approach significantly outperforms the baselines across all three datasets in terms of MAP and Spearman's correlation measures, demonstrating its effectiveness.
Further quantitative experiments show that our approach achieves better performance when leveraging more dialogue context and remains robust when less training data is provided.
\end{abstract}

\section{Introduction}
Sentence embeddings are used with success for a variety of NLP applications \cite{use} and many prior methods have been proposed with different learning schemes.
\citet{skip_thought, quick_thought, fast_sent} train sentence encoders in a self-supervised manner with web pages and books.
\citet{infer_sent, use, sent_bert} propose to learn sentence embeddings on the supervised datasets such as SNLI \cite{snli} and MNLI \cite{mnli}.
Although the supervised-learning approaches achieve better performance, they suffer from high cost of annotation in building the training dataset, which makes them hard to adapt to other domains or languages.

Recently, learning sentence embeddings from dialogues has begun to attract increasing attention.
Dialogues provide strong semantic relationships among conversational utterances and are usually easy to collect in large amounts.
Such advantages make the dialogue-based self-supervised learning methods promising to achieve competitive or even superior performance against the supervised-learning methods, especially under the low-resource conditions.

While promising, the issue of how to effectively exploit the dialogues for this task has not been sufficiently explored.
\citet{co_dan} propose to train an input-response prediction model on Reddit dataset \cite{reddit}.
Since they build their architecture based on the single-turn dialogue, the multi-turn dialogue history is not fully exploited.
\citet{convert} demonstrate that introducing the multi-turn dialogue context can improve the sentence embedding performance.
However, they concatenate the multi-turn dialogue context into a long token sequence, failing to model inter-sentence semantic relationships among the utterances.
Recently, more advanced methods such as \cite{sent_bert} achieve better performance by employing BERT \cite{bert} as the sentence encoder.
These works have in common that they employ a feed-forward network with a non-linear activation on top of the sentence encoders to model the context-response semantic relevance, thereby learning the sentence embeddings.
However, such architecture presents two limitations: (1) It yields a large gap between training and evaluating, since the semantic textual similarity is commonly measured by the element-wise distance metrics such as cosine and L2 distance.
(2) Concatenating all the utterances in the dialogue context inevitably introduces the noise as well as the redundant information, resulting in a poor result.

In this paper, we propose DialogueCSE, a dialogue-based contrastive learning approach to tackle these issues.
We hold that the semantic matching relationships between the context and the response can be implicitly modeled through contrastive learning, thus making it possible to eliminate the gap between training and evaluating.
To this end, we introduce a novel matching-guided embedding (MGE) mechanism.
Specifically, MGE first pairs each utterance in the context with the response and performs a token-level dot-product operation across all the utterance-response pairs to obtain the multi-turn matching matrices.
Then the multi-turn matching matrices are used as guidance to generate a context-aware embedding for the response embedding (i.e. the context-free embedding).
Finally, the context-aware embedding and the context-free embedding are paired as a training sample, whose label is determined by whether the context and the response are originally from the same dialogue.
Our motivation is that once the context semantically matches the response, it has the ability to distill the context-aware information from the context-free embedding, which is exactly the learning objective of the sentence encoder that aims to produce context-aware sentence embeddings.

We train our model on three multi-turn dialogue datasets: the Microsoft Dialogue Corpus (MDC) \cite{mdc}, the Jing Dong Dialogue Corpus (JDDC) \cite{jddc}, and the E-commerce Dialogue Corpus (ECD) \cite{edc}.
To evaluate our model, we introduce two types of tasks: the semantic retrieval (SR) task and the dialogue-based semantic textual similarity (D-STS) task.
Here we do not adopt the standard semantic textual similarity (STS) task \cite{sts_b} for two reasons:
(1) As revealed in \cite{is_bert}, the sentence embedding performance varies greatly as the domain of the training data changes.
As a dialogue dataset is always about several certain domains, evaluating on the STS benchmark may mislead the evaluation of the model.
(2) The dialogue-based sentence embeddings focus on context-aware rather than context-free semantic meanings, which may not be suitable to be evaluated through the context-free benchmarks.
Since previous dialogue-based works have not set up a uniform benchmark, we construct two evaluation datasets for each dialogue corpus.
A total of 18,964 retrieval samples and 4,000 sentence pairs are annotated by seven native speakers through the crowd-sourcing platform\footnote{All the datasets will be publicly available at https://github.com/wangruicn/DialogueCSE}.
The evaluation results indicate that DialogueCSE significantly outperforms the baselines on the three datasets in terms of both MAP and Spearman's correlation metrics, demonstrating its effectiveness.
Further quantitative experiments show that DialogueCSE achieves better performance when leveraging more dialogue context and remains robust when less training data is provided.
To sum up, our contributions are threefold:
\begin{itemize}
  \item We propose DialogueCSE, a dialogue-based contrastive learning approach with MGE mechanism for learning sentence embeddings from dialogues. 
  As far as we know, this is the first attempt to apply contrastive learning in this area.
  \item We construct the dialogue-based sentence embedding evaluation benchmarks for three dialogue corpus. 
  All of the datasets will be released to facilitate the follow-up researches.
  \item Extensive experiments show that DialogueCSE significantly outperforms the baselines, establishing the state-of-the-art results.
\end{itemize}

\section{Related Work}
\subsection{Self-supervised Learning Approaches}
Early works on sentence embeddings mainly focus on the self-supervised learning approaches.
\citet{skip_thought} train a seq2seq network by decoding the token-level sequences of the context in the corpus.
\citet{fast_sent} propose to predict the neighboring sentences as bag-of-words instead of step-by-step decoding.
\citet{quick_thought} perform sentence-level modeling by retrieving the ground-truth sentence from candidates under the given context, achieving consistently better performance compared to the previous token-level modeling approaches.
The datasets used in these works are typically built upon the corpus of web pages and books \cite{toronto}.
As the semantic connections are relatively weak in these corpora, the model performances in these works are inherently limited and hard to achieve further improvement.

Recently, the pre-trained language models such as BERT \cite{bert} and GPT \cite{gpt_2} yield strong performances across many downstream tasks \cite{wang2018glue}.
However, BERT's embeddings show poor performance without fine-tuning and many efforts have been devoted to alleviating this issue.
\citet{is_bert} propose a self-supervised learning approach that derives meaningful BERT sentence embeddings by maximizing the mutual information between the global sentence embedding and all its local context embeddings.
\citet{bert_flow} argue that BERT induces a non-smooth anisotropic semantic space.
They propose to use a flow-based generative module to transform BERT's embeddings into isotropic semantic space.
Similar to this work, \citet{bert_whitening} replace the flow-based generative module with a simple but efficient linear mapping layer, achieving competitive results with reported experiments in BERT-flow.

Lately, the contrastive self-supervised learning approaches have shown their effectiveness and merit in this area.
\citet{wu2020clear, giorgi2020declutr, meng2021coco} incorporate the data augmentation methods including the word-level deletion, reordering, substitution, and the sentence-level corruption into the pre-training of deep Transformer models to improve the sentence representation ability, achieving significantly better performance than BERT especially on the sentence-level tasks \cite{wang2018glue, sts_b, conneau2018senteval}.
\citet{gao2021simcse} apply a twice independent dropout to obtain two same-source embeddings from a single sentence as input.
Through optimizing their cosine distance, SimCSE achieves remarkable gains over the previous baselines.
\citet{yan2021consert} empirically study more data augmentation strategies in learning sentence embeddings, and it also achieves remarkable performance as SimCSE.
In this work, we propose the MGE mechanism to generate a context-aware embedding for each candidate response based on its context-free embedding.
Different from previous methods built upon the data augmentation strategies, MGE leverages the context to accomplish this goal without any text corruption.

For dialogue, \citet{co_dan} train a siamese transformer network with single-turn input-response pairs extracted from Reddit.
Such architecture is further extended in \cite{sent_bert} by replacing the transformer encoder with BERT.
\citet{convert} propose to leverage the dialogue context to improve the sentence embedding performance.
They concatenate the multi-turn dialogue context into a long word sequence and adopt a similar architecture as \cite{co_dan} to model the context-response matching relationships.
Our work is closely related to their works.
We propose a novel dialogue-based contrastive learning approach, which directly models the context-response matching relationships without an intermediate MLP.
We also consider the interactions between each utterance in the dialogue context and the response instead of simply treating the dialogue context as a long sequence.

\subsection{Supervised Learning Approaches}
The supervised learning approaches mainly focus on training classification models with the SNLI and the MNLI datasets \cite{snli, mnli}.
\citet{infer_sent} demonstrate the superior performance of the supervised learning model on both the STS-benchmark \cite{sts_b} and the SICK-R tasks \cite{sick_r}.
Based on this observation, \citet{use} further extend the supervised learning to the multi-task learning by introducing the QA prediction task, the Skip-Thought-like task \cite{henderson2017efficient,skip_thought}, and the NLI classification task, achieving significant improvement over InferSent.
\citet{sent_bert} employ BERT as sentence encoders in the siamese-network and fine-tune them with the SNLI and the MNLI datasets, achieving the new state-of-the-art performance.

\begin{figure*}[thb]
  \centering
  \includegraphics[width=\linewidth]{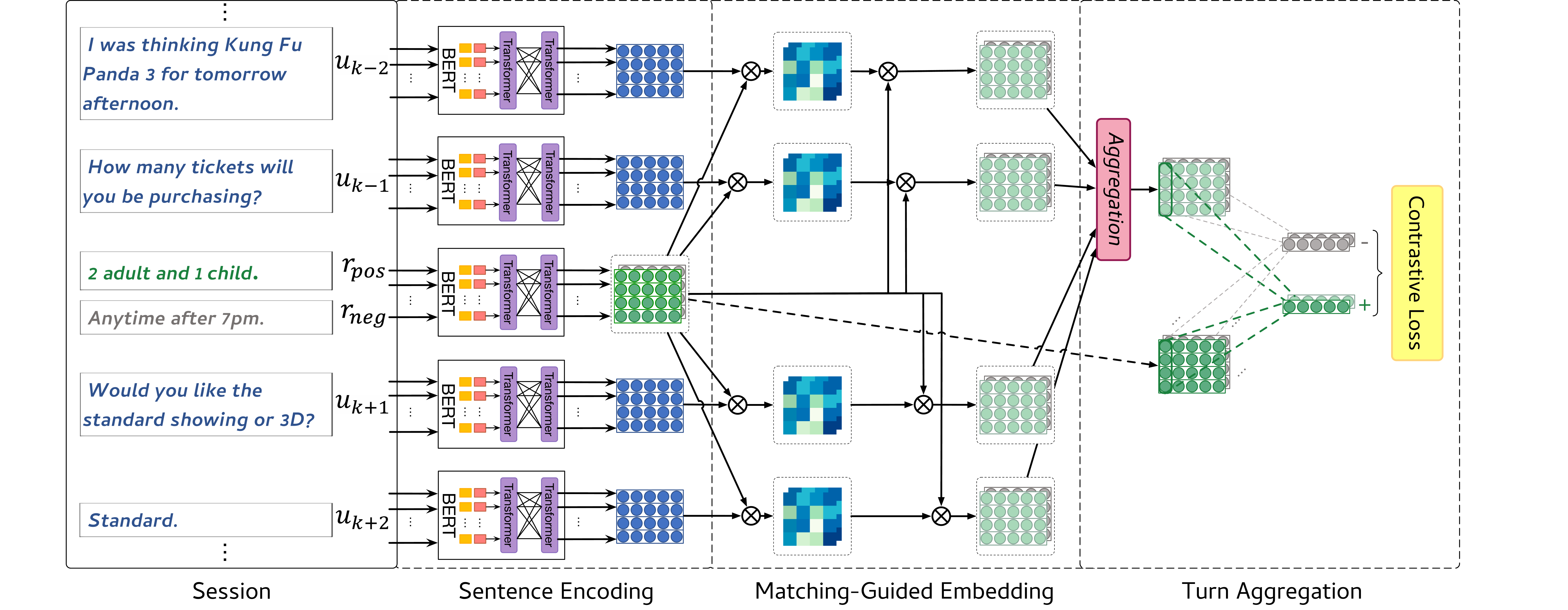}
  \caption{Model architecture. (1) We use BERT to encode the multi-turn dialogue context and the responses, all of the BERT encoders share the same parameters. (2) The matching-guided embedding (MGE) mechanism performs the token-level matching between each utterance and a response, generates multiple refined embeddings across turns. (3) All refined embedding matrices are aggregated to form a context-aware embedding matrix, which is further pooled along the sequence dimension.}
  \label{figure:model}
\end{figure*}

\vspace{1cm}

\section{Problem Formulation}
Suppose that we have a dialogue dataset $\mathcal{D}=\{ S_{i} \}_{i=1}^{K}$, where $S_{i}=\{u_{1}, \cdots, u_{k-1}, r, u_{k+1}, \cdots, u_{t}\}$ is the $i$-th dialogue session in $\mathcal{D}$ with $t$ turn utterances.
$r$ is the response and $C_{i}=\{u_{1}, \cdots\, u_{k-1}, u_{k+1}, \cdots, u_{t}\}$ is the bi-directional context around $r$.
We omit the subscript $i$ in the following paragraph and use $S, C$ instead of $S_{i}, C_{i}$ for brevity.

To generate the contrastive training pairs, we introduce two embedding matrices for $r$, named context-free embedding matrix and context-aware embedding matrix.
Specifically, we first encode $r$ as an embedding matrix $\Bar{\textbf{R}}$.
Since $\Bar{\textbf{R}}$ is encoded independently of the dialogue context, it is treated as the context-free embedding matrix.
Then we generate a corresponding embedding matrix $\tilde{\textbf{R}}$ based on $\Bar{\textbf{R}}$ according to the guidance of $C$.
$\tilde{\textbf{R}}$ is treated as the context-aware embedding matrix.
As $C$ and $r$ are derived from the same dialogue, ($\Bar{\textbf{R}}, \tilde{\textbf{R}}$) naturally forms a positive training pair.
To construct a negative training pair, we first sample an utterance $r'$ from a dialogue randomly selected from $\mathcal{D}$.
$r'$ is encoded as the context-free embedding matrix $\Bar{\textbf{R}}'$ based on which a context-aware embedding matrix $\tilde{\textbf{R}}'$ is generated through the completely identical process.
($\Bar{\textbf{R}}', \tilde{\textbf{R}}'$) is treated as a negative training pair.
For each response $r$, we generate a positive training pair (since there is only one ground-truth response for each context) and multiple negative training pairs.
All the training pairs are then passed through the contrastive learning module.

It is worth to mention that there is no difference between sampling the response or the context as they are symmetrical in constructing the negative training pairs.
But we prefer the former as it is more straightforward and in accordance with the previous retrieval-based works for dialogues.
With all the training samples at hand, our goal is to minimize their contrastive loss, thus fine-tuning BERT as a context-aware sentence encoder.

\section{Our Approach}
Figure \ref{figure:model} shows the model architecture. 
Our model is divided into three stages: sentence encoding, matching-guided embedding, and turn aggregation.
We describe each part as below.

\subsection{Sentence Encoding}
We adopt BERT \cite{bert} as the sentence encoder.
Let $u$ represent a certain utterance in $C$.
$u$ and $r$ are first encoded as two sequences of output embeddings, which is formulated as:
\begin{align}
    & \left \{\textbf{u}_{1}, \textbf{u}_{2}, \cdots, \textbf{u}_{n} \right \} = \textbf{BERT}(u), \\
    & \left \{\textbf{r}_{1}, \textbf{r}_{2}, \cdots, \textbf{r}_{n} \right \} = \textbf{BERT}(r),
\end{align}
where $\textbf{u}_{i}$, $\textbf{r}_{j}$ represent the $i$-th and the $j$-th output embedding derived from $u$ and $r$ respectively.
$n$ is the maximum sequence length of both input sentences.
$\forall{i, j} \in {1, 2, \cdots, n}$, the shapes of $\textbf{u}_{i}$ and $\textbf{r}_{j}$ are $1 \times d$, where $d$ is the dimension of BERT's outputs.
We stack $\left \{\textbf{u}_{1}, \textbf{u}_{2}, \cdots, \textbf{u}_{n}\right \}$ and $\left \{\textbf{r}_{1}, \textbf{r}_{2}, \cdots, \textbf{r}_{n}\right \}$ to obtain the context-free embedding matrices $\Bar{\textbf{U}}$ and $\Bar{\textbf{R}}$, whose shapes are both $n \times d$.

\subsection{Matching-Guided Embedding}
The matching-guided embedding mechanism performs a token-level matching operation on $\Bar{\textbf{U}}$ and $\Bar{\textbf{R}}$ to form a matching matrix $\textbf{M}$, which is formulated as:
\begin{align}
    \textbf{M} = \frac{\Bar{\textbf{U}} \left ( \Bar{\textbf{R}} \right )^\mathrm{T}}{\sqrt{d}},
\end{align}
Then it generates a refined embedding matrix $\hat{\textbf{R}}$ based on the context-free embedding matrix $\Bar{\textbf{R}}$, which is given by:
\begin{align}
    \hat{\textbf{R}} = \textbf{M} \Bar{\textbf{R}}
\end{align}
$\hat{\textbf{R}}$ is a new representation of $r$ from the perspective of the utterance $u$.
Note that as $u$ is only a single turn utterance in $C$, we generate $t-1$ refined embedding matrices for $r$ in total.

\subsection{Turn Aggregation}
After obtaining all of the refined embedding matrices across turns, we consider two strategies to fuse them to obtain the final context-aware embedding matrix $\tilde{\textbf{R}}$.
The first strategy adopts a weighted sum operation based on the attention mechanism, formulated by:
\begin{align}
  \tilde{\textbf{R}} = \sum_{i} \alpha_{i} \hat{\textbf{R}}_{i},
\end{align}

where $i \in \left \{1, \cdots, k-1, k+1, \cdots, t \right \}$ and $\hat{\textbf{R}}_{i}$ is the refined embedding matrix corresponding to the $i$-th turn utterance in the context. 
The attention weight $\alpha_{i}$ is decided by:
\begin{align}
  \label{align:weighted pooling operation}
  \alpha_{i} = \frac{\exp(\text{FFN}(\hat{\textbf{R}}_{i}))}{\sum_{j}\exp( \text{FFN}(\hat{\textbf{R}}_{j}))},
\end{align}
where $\text{FFN}$ is a two-layer feed-forward network with \text{ReLU} \cite{relu} activation function.
We denote this strategy as $I_{1}$.
The second strategy $I_{2}$ directly sums up all the refined embeddings across turns, which is defined as:
\begin{align}
  \tilde{\textbf{R}} = \dfrac{1}{t-1}\sum_{i} \hat{\textbf{R}}_{i},
\end{align}

For the negative sample $r'$, we apply the same procedure to generate the context-free embedding matrix $\Bar{\textbf{R}}'$ and the context-aware embedding $\tilde{\textbf{R}}'$.
Each context-aware embedding matrix is then paired with its corresponding context-free embedding matrix to form a training pair.

As mentioned in the introduction, MGE holds several advantages in modeling the context-response semantic relationships.
Firstly, the token-level matching operation acts as a guide to distill the context-aware information from the context-free embedding matrix.
Meanwhile, it provides rich semantic matching information to assist the generation of the context-aware embedding matrix.
Secondly, MGE is lightweight and computationally efficient, which makes the model easier to train than the siamese-network-based models.
Finally and most importantly, the context-aware embedding $\tilde{\textbf{R}}$ shares the same semantic space with $\Bar{\textbf{R}}$, which enables us to directly measure their cosine similarity.
This is the key to successfully model the semantic matching relationships between the context and the response through contrastive learning.

\subsection{Learning Objective}
We adopt the NT-Xent loss proposed in \cite{oord2018representation} to train our model.
The loss $\mathcal{L}$ is formulated as:
\begin{equation}
    \begin{aligned}
        \mathcal{L} = -\frac{1}{N}\sum_{i=1}^{N} \log\frac{ e^{\text{sim}(\Bar{\textbf{R}}_{i}, \tilde{\textbf{R}}_{i})/\tau} }{\sum_{j=1}^{M}e^{\text{sim}(\Bar{\textbf{R}}_{j}, \tilde{\textbf{R}}_{j})/\tau }},
    \end{aligned}
\end{equation}
where $N$ is the number of all the positive training samples and $M$ is the number of all the training pairs associated with each positive training sample $r$.
$\tau$ is the temperature hyper-parameter.
$\text{sim}(\cdot,\cdot)$ is the similarity function, defined as a token-level pooling operation followed by the cosine similarity.

Once the model is trained, we take the mean pooling of BERT's output embeddings as the sentence embedding.

\section{Experiments}
We conduct experiments on three multi-turn dialogue datasets: the Microsoft Dialogue Corpus (MDC) \cite{mdc}, the Jing Dong Dialogue Corpus (JDDC) \cite{jddc}, and the E-commerce Dialogue Corpus (ECD) \cite{edc}.
Each utterance in these three datasets is originally assigned with an intent label, which is further leveraged by us in the heuristic strategy to construct the evaluation datasets.

\subsection{Experimental Setup}
\subsubsection{Training}
Table \ref{tab:dataset_profile} shows the statistics information of these three datasets.
The Microsoft Dialogue Corpus is a task-oriented dialogue dataset.
It consists of three domains, each with 11 identical intents.
The Jing Dong Dialogue Corpus is a large-scale customer service dialogue dataset publicly available from JD\footnote{https://www.jd.com}.
Although the dataset collected from the real-world scenario is quite large, it contains much noise which brings great challenges for our model.
The E-commerce Dialogue Corpus is a large-scale dialogue dataset collected from Taobao\footnote{https://www.taobao.com}.
The released dataset takes the form of the response selection task.
We recover it to the dialogue sessions by dropping the negative samples and splitting the context into multiple utterances.
We pre-process these datasets by the following steps: (1) We combine the consecutive utterances of the same speaker. (2) We discard the dialogues with less than 4 turns in JDDC and ECD since such dialogues are usually incomplete in practice.

\begin{table}[tb]
    \centering
    \small
    \begin{tabular}{lccccc}
        \toprule
        \textbf{Dataset} & \textbf{MDC} & \textbf{JDDC} & \textbf{ECD}\\
        \midrule
        \# Total dialogues    &  10,087   &   1,024,196   &   1,020,000       \\
        \# Total turns   &  74,685   &   20,451,337  &   7,500,000   \\
        \# Total words        &  190,952         &   150,716,172 &   49,000,000 \\
        \# Total intents      &  11       &   289         &   207  \\
        \bottomrule
    \end{tabular}
    \caption{Statistics of the datasets.}
    \label{tab:dataset_profile}
\end{table}

\begin{table*}[htb]
    \centering
    \small
    \begin{tabular}{l|ccc|ccc|ccc}
        \toprule

        \multirow{2}{*}{\textbf{ Model}} 
        & \multicolumn{3}{c}{\textbf{Microsoft Corpus}} 
        & \multicolumn{3}{|c}{\textbf{Jing Dong Corpus}}
        & \multicolumn{3}{|c}{\textbf{E-commerce Corpus}}\\

        & Corr. & MAP & MRR  
        & Corr. & MAP & MRR
        & Corr. & MAP & MRR   \\
        
        \midrule
        \midrule
        \multicolumn{10}{c}{\textit{Self-supervised models}} \\
        \midrule

        Avg. GloVe embeddings & 36.64 & 31.59 & 40.91 & 39.61 & 45.94 & 59.53 & 19.80 & 46.14 & 63.68  \\
        BERT-CLS & 22.34 & 29.54 & 35.94 & 21.40 & 45.05 & 59.58 & 16.61 & 47.75 & 65.91   \\
        BERT-avg & 40.95 & 32.10 & 43.01  & 50.89 & 49.08 & 64.54  & 43.68 & 51.77 & 70.79   \\
        BERT-flow & 45.56 & 33.13 & 40.86  & 65.11 & 49.53 & 64.30 & 55.04  & 52.16 & 71.06   \\
        BERT-whitening & 26.70 & 32.09 & 43.01 & 61.57 & 49.08 & 64.54 & 47.64 & 51.77 & 70.80 \\
        
        \midrule
        
        BERT(adapt)-CLS & 27.35 & 31.30 & 39.83 & 26.49 & 48.51 & 65.70 & 33.91 & 51.75 &  74.68 \\
        BERT(adapt)-avg  & 42.81 & 32.53 & 43.49  & 72.60 & 53.03 & 66.99 & 74.26 & 59.32 & 76.89   \\
        BERT(adapt)-flow  & 50.17 & 34.32 & 41.62 & 73.32 & 53.42 & 67.00 & 74.31 & 59.77 & 76.48 \\
        BERT(adapt)-whitening   & 29.68 & 32.53 & 43.48 & 67.18 & 53.04 & 67.01 & 57.22 & 59.33 & 76.84 \\
        
        \midrule
        \midrule
        \multicolumn{10}{c}{\textit{Dialogue-based self-supervised models}} \\
        \midrule
        
        SiameseBERT$_{S}$ & 77.95 & 76.26 & 84.92 & 75.70 & 61.92 & 74.44 & 74.83 & 65.84 & 79.88  \\
        SiameseBERT$_{M}$ & 76.70 & 73.81 & 85.09 & 76.85 & 62.45 & 74.64 & 75.45 & 66.24 & 80.58  \\
        \midrule
        
        DialogueCSE$_{I_{1}}$ & 80.13	& 87.26 & 85.89 & 80.60 & 66.54 & 74.79 & 81.79 & 68.70 & 79.89  \\
        
        \textbf{DialogueCSE$_{I_{2}}$}
        & \textbf{82.36} & \textbf{91.40} & \textbf{90.45} 
        & \textbf{81.22} & \textbf{68.02} & \textbf{79.52} 
        & \textbf{83.94} & \textbf{69.32} & \textbf{81.20}  \\
        
        \bottomrule
    \end{tabular}
    \caption{Evaluation results on the dialogue-based semantic textual similarity (D-STS) task and the semantic retrieval (SR) task. Corr. refers to Spearman's correlation metric for the D-STS task. MAP and MRR are metrics for the SR task. Reported numbers are in percentages.}
    \label{main experiment result}
\end{table*}

\subsubsection{Evaluation}
We introduce the semantic retrieval (SR) and the dialogue-based STS (D-STS) tasks to evaluate our model.
For the SR task, we construct evaluation datasets by the following steps:
(1) we sample a large number of sentences with the intent labels as candidates.
(2) the candidates are annotated with binary labels indicating whether the given sentence and its intent label are consistent.
The inconsistent instances are directly discarded from the candidates.
(3) for each sentence, we retrieval 100 sentences through BM25 \cite{bm25} from the candidates, and assign each candidate sentence a label by whether its intent is consistent with the target sentence.
We limit the number of positive samples to a maximum of 30 and keep approximately 7k, 7k, and 4k samples for MDC, JDDC, and ECD respectively.

For the D-STS task, we sample the sentence pairs from the dialogues following the heuristic strategies proposed by \cite{sts_b} to ensure there are enough semantically similar samples.
The heuristic strategies include unigram-based and w2v-based KNN retrieval methods and random sampling from the candidates with the same intent labels.
The sentence pairs are further annotated through the crowd-sourcing platform, with five degrees ranging from 1 to 5 according to their semantic relevance.
We use the median number of annotated results as the semantic relevance degrees, obtaining 1k, 2k, and 1k sentence pairs for MDC, JDDC, and ECD respectively.

All annotations are carried out by seven native speakers.
For the SR task, we adopt the Mean average precision (MAP) and the Mean reciprocal rank (MRR) metrics.
Following previous works, we adopt Spearman's correlation metric for the D-STS task to assess the quality of the dialogue-based sentence embeddings.

\subsection{Baselines}
We evaluate our model against the two groups of baselines: self-supervised learning methods and dialogue-based self-supervised learning methods.
The former is not designed for dialogues while the latter is.

\subsubsection{\textbf{Self-supervised learning methods}}
In this line, we consider the BERT-based methods, which include BERT \cite{bert}, domain-adaptive BERT \cite{bert_dapt}, BERT-flow \cite{bert_flow}, and BERT-whitening \cite{bert_whitening}.
"Domain-adaptive BERT" means that we run continue pre-training with the dialogue datasets.
BERT-flow and BERT-whitening are two BERT-based variants that transform BERT's sentence embedding to the isotropic semantic space.

For BERT, we use the \texttt{[CLS]} token embedding (denoted as BERT-CLS) and the average of the sequence output embeddings (denoted as BERT-avg) as the sentence embedding, and the same is true for domain-adaptive BERT.
It should be noted that in related sentence embedding researches, domain-adaptive BERT is rarely considered since the training datasets are relatively small.
Fortunately, the large-scale dialogue datasets allow us to explore whether the domain-adaptive pre-training is helpful for our tasks.
We also adopt the average of GloVe word embeddings \cite{pennington2014glove} (denoted as Avg. GloVe) as the sentence embedding to compare with our results.

\subsubsection{\textbf{Dialogue-based self-supervised learning methods}}
In this line, we mainly consider the siamese-networks commonly applied in dialogue-based researches.
Considering none of the previous works \cite{co_dan, convert} employs the pre-trained language model as encoder, we re-implement two BERT-based siamese-network models according to their original approaches.
The first baseline $\text{SiameseBERT}_{s}$ is a siamese-network which shares the architecture with \cite{co_dan, sent_bert}.
It is equipped with a non-linear activation function in the matching layer to model the heterogeneous matching relationships between the context and the response\footnote{We use "heterogeneous" to describe the matching relationships for context-response pairs since they have different semantic meanings. As a comparison, the NIL-like sentence pairs have the "homogeneous" matching relationships.}.
The second baseline $\text{SiameseBERT}_{m}$ has the similar architecture as \cite{convert}.
It flattens the multi-turn context and takes the token sequence as input.
There is also an MLP layer on top of the sentence encoders.

\subsection{Implementation Details}
Our approach is implemented in Tensorflow \cite{abadi2016tensorflow} with CUDA 10.0 support.
For all datasets, we continue pre-training BERT for approximately 0.5 epochs to improve its domain adaption ability as well as keeping the general domain information as much as possible.
During the continue pre-training stage, we use a masking probability of 0.15, a learning rate of 2e-5, a batch size of 50, and a maximum of 10 masked LM predictions per sequence.
During the contrastive learning stage, we freeze the bottom 6 layers of BERT to prevent catastrophic forgetting which simultaneously enables the model to be trained with larger batch size.
Such a setting achieves the best performance in our experiments.
The batch size, the learning rate, and the number of context turns are set to 20, 5e-5, and 3 respectively.
The maximum sequence length is set to 100, 50, 50 for JDDC, MDC, and ECD for both continue pre-training stage and contrastive learning stage.
All models are trained on 4 Tesla V100 GPUs.

\subsection{Evaluation Results} 
Table \ref{main experiment result} shows the main experimental results on the three datasets.
From the table, we can observe that our model achieves the best performance in terms of all metrics across the three datasets.
Compared to the results of the siamese-networks, our model achieves at least 4.41 points (77.95 $\rightarrow$ 82.36),  4.37 points (76.85 $\rightarrow$ 81.22), and 8.49 points (75.45 $\rightarrow$ 83.94) in terms of Spearman's correlation on MDC, JDDC, and ECD respectively.
It also improves the MAP metric by 14.84 points (76.26 $\rightarrow$ 91.40), 5.57 points (62.45 $\rightarrow$ 68.02), and 3.08 points (66.24 $\rightarrow$ 69.32) in terms of MAP metric on the three datasets.
There are even larger improvements between DialogueCSE and the domain-adaptive baselines including BERT(adapt) and its variants.
We attribute this improvement to two main reasons: First, by introducing contrastive learning, DialogueCSE eliminates the gap between training and evaluating, gaining significant improvements on both SR and D-STS tasks.
Second, DialogueCSE models the semantic relationships in each utterance-response pair, which distills the important information at turn-level from the multi-turn dialogue context and achieves better performance. 

Moreover, by comparing the performances of $\text{DialogueCSE}_{I_{1}}$ and $\text{DialogueCSE}_{I_{2}}$, we find that the weighted sum aggregation strategy surprisingly brings a significant deterioration on all metrics.
We consider that this is because the weighted sum operation breaks down the turn-level unbiased aggregation process.
Since the attention mechanism tends to provide shortcuts for the model to achieve its learning objective, the long-tail utterances in the context may be partially ignored, thus leading to a decline in embedding performance.
We hold that we can completely dismiss the weighted sum aggregation strategy in DialogueCSE since the token-level matching operation in MGE has implicitly served this role.

We also notice that BERT(adapt) achieves significantly better performance than the original BERT, especially on JDDC and ECD.
It demonstrates the importance of continued pre-training with the in-domain training data.
Without such procedure, the in-domain data can't be fully exploited, making it difficult for the model to achieve satisfactory performance.
This also indicates that the MLM pre-training task is indeed a powerful task to learn effective sentence embeddings from texts, especially when the domain training data is sufficient.

\subsection{Discussion}
We conduct comparison and hyper-parameter experiments in the following section to study how our model performs with different numbers of turns, data scales, temperature hyper-parameter, and numbers of negative samples.

\subsubsection{Comparison with Baseline}
In this section, we choose $\text{SiameseBERT}_{m}$ as a comparison method.
MAP and Spearman's correlation metrics are adopted in these experiments.

\begin{figure}[htb]
  \centering
  \includegraphics[width=\linewidth]{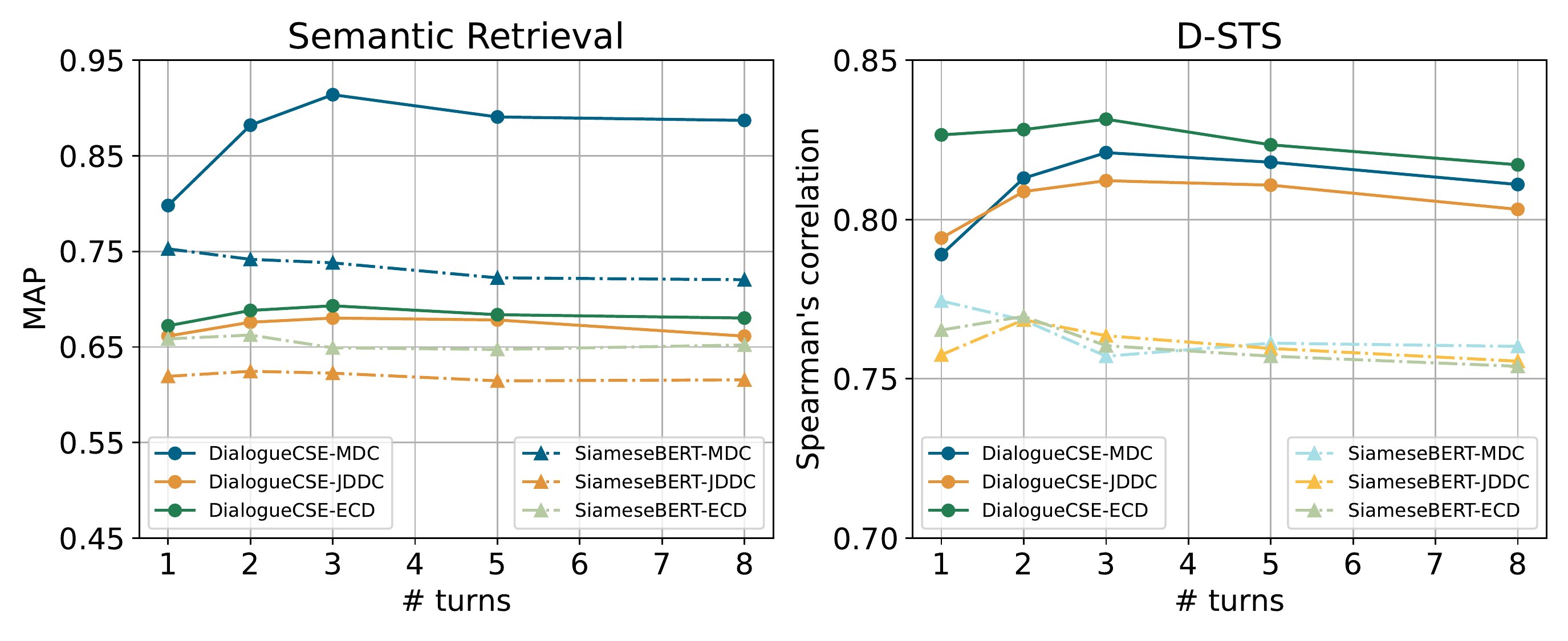}
  \caption{Impact of turn number.}
  \label{figure:turn}
\end{figure}

{\textbf{Impact of turn number}}. 
Figure \ref{figure:turn} shows the performance of our model and the baseline under different numbers of turns on all datasets.
From the results, we observe that our model is indeed benefited from the multi-turn dialogue context, and it exhibits consistently better performance than the baseline.
The performance of our model increases as the turn number increases until it approximately arrives at 3.
When the turn number goes bigger, the performance of both models begins to drop.
We believe that in this case, adding more dialogue context will bring too much noise.
Since MGE acts as a noise filter at both token and turn level, it makes the model more robust when using more context turns.

\begin{figure}[htb]
  \centering
  \includegraphics[width=\linewidth]{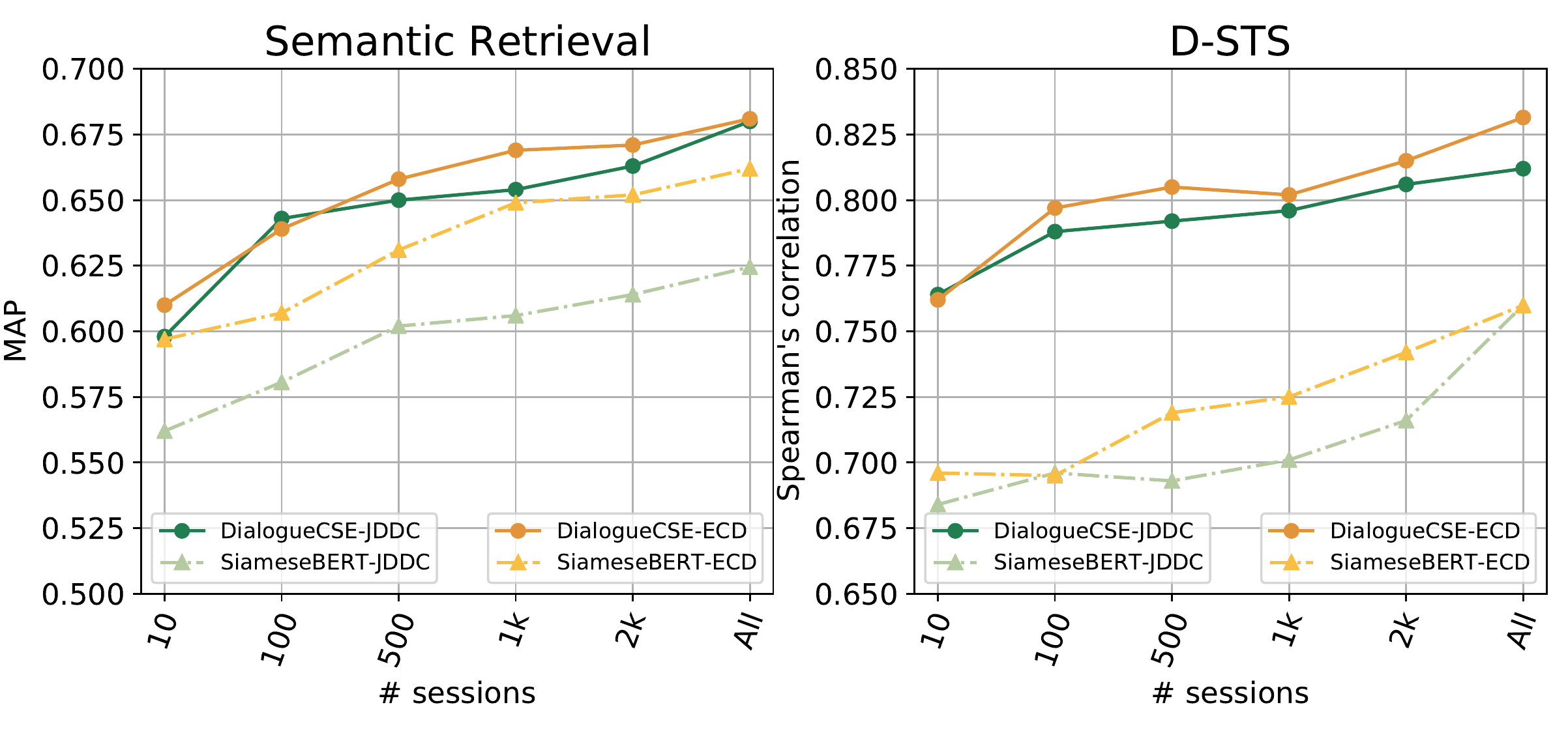}
  \caption{Impact of data scale.} 
  \label{figure:session}
\end{figure}

{\textbf{Impact of data scale}}. 
We further explore whether our model is robust when fewer training samples are given.
We select JDDC and ECD in this experiment since they are large-scale and topically diverse, which is suitable for simulating a few-shot learning scenario.
Figure \ref{figure:session} shows the performances of our model and the baseline under different numbers of training dialogues.

As the figure reveals, the performance gaps between our model and the baseline are even larger when fewer training dialogue sessions are given.
Particularly, when using only a few dialogues, our model can achieve even superior performance over the $\text{SiameseBERT}$ trained on larger datasets, especially on the D-STS task.
We think this is reasonable since the siamese-networks introduce a large amount parameters to model the semantic matching relationships, while our model accomplishes this goal without introducing any additional parameters.

\subsubsection{Hyper-parameter Evaluations}
We further conduct experiments on JDDC and EDC to study how our model is influenced by the temperature $\tau$ and the number of negative samples.
The MDC dataset is excluded here since the semantics of its utterances are highly centralized around a few top intents.

{\textbf{Impact of temperature}}.
Table \ref{tab:hyper-parameter} shows the experimental results with different $\tau$ values.
We find that the Spearman's correlations increase monotonically as $\tau$ increases until $0.1$ for JDDC and $0.2$ for ECD, then they begin to drop.
The MAP metrics also increase as $\tau$ increases until $0.1$ for both datasets, but they remain stable as $\tau$ varies from $0.1$ to $0.5$.
We consider this is due to the coarse-grained nature of the SR task.
When $\tau$ approaches $0.1$, our model can gradually distinguish among different fine-grained semantics, thus achieving better performance on both SR and D-STS tasks.
As $\tau$ continues to increase, the model forces the sentence embeddings to be closer, resulting in a decrease in Spearman's correlation.
However, as all positive samples in the candidates have identical labels, such degradation may not be fully reflected through the ranking metric (e.g. MAP) or even be covered as the number of retrieved positive samples changes.

\begin{table}[tb]
    \centering
    \small
    \begin{tabular}{lc|cccc}
        \toprule
        \multicolumn{2}{c|}{\textbf{Temperature}} & \textbf{0.05} & \textbf{0.1} & \textbf{0.2}  & \textbf{0.5} \\
        \midrule
        \multirow{2}{*}{\textbf{JDDC}} & Corr. & 80.05  &  81.22  &  80.82  &  79.85   \\
                                       & MAP   & 67.19  &  68.02  &  67.55  &  68.63    \\
        \midrule
        \multirow{2}{*}{\textbf{ECD}} & Corr.  & 82.24  &  83.94  &  84.24  &  83.76       \\
                                      & MAP    & 67.98  &  69.32  &  69.63  &  69.11       \\
        \bottomrule
    \end{tabular}
    \caption{Impact of temperature.}
    \label{tab:hyper-parameter}
\end{table}

{\textbf{Impact of negative samples}}.
We vary the number of negative samples for each positive sample within $\{\text{1, 4, 9, 19}\}$.
Table \ref{tab:negative samples} shows the experimental results, from which we find that both metrics improve slightly when the number of negative samples increases.
Considering the similar observation in \cite{gao2021simcse, yan2021consert}, we conclude this phenomenon may be related to the discrete nature of language.
Specifically, as the generation of the sentence embeddings in our approach is guided and constrained by the token-level interaction mechanism, our model is more robust than the other contrastive learning approaches and is even effective when only one negative sample is provided.

\section{Conclusion}
In this work, we propose DialogueCSE, a dialogue-based contrastive learning approach to learn sentence embeddings from dialogues.
We also propose uniform evaluation benchmarks for evaluating the quality of the dialogue-based sentence embeddings. 
Evaluation results show that DialogueCSE achieves the best result over the baselines while adding no additional parameters.
In the next step, we will study how to introduce more interaction information to learn the sentence embeddings and try to incorporate the contrast learning method into the pre-training stage.

\begin{table}[tb]
    \centering
    \small
    \begin{tabular}{lc|ccccc}
        \toprule
        \multicolumn{2}{c|}{\textbf{\# Negative samples}} & \textbf{1} & \textbf{4} & \textbf{9}  & \textbf{19} \\
        \midrule
        \multirow{2}{*}{\textbf{JDDC}} & Corr. & 80.60  &  80.85  &  81.22  &  81.56  \\
                                       & MAP   & 67.48  &  67.69  &  68.02  &  68.63     \\
        \midrule
        \multirow{2}{*}{\textbf{ECD}} & Corr.   & 82.55   &   83.14   &   83.94 & 84.12     \\
                                       & MAP   & 68.56   &   68.87   &   69.32 & 69.56    \\
        \bottomrule
    \end{tabular}
    \caption{Impact of negative samples.}
    \label{tab:negative samples}
\end{table}


\bibliography{anthology,custom}
\bibliographystyle{acl_natbib}




\end{document}